\documentclass[letterpaper, 10 pt, conference]{ieeeconf}  

\IEEEoverridecommandlockouts                              

\usepackage{graphics} 
\usepackage{epsfig} 
\usepackage{mathptmx} 
\usepackage{times} 
\usepackage{amsmath} 
\usepackage{amssymb}  
\usepackage{caption}
\usepackage{subcaption}

\usepackage{todonotes}

\title{\LARGE \bf
Enabling and Assessing Trust when Cooperating with Robots in Disaster Response (EASIER)*
}

\author{Laurent Frering$^{1}$, Matthias Eder$^{1}$, Bettina Kubicek$^{2}$ Dietrich Albert$^{2}$, Denis Kalkofen$^{3}$, Thomas Gschwandtner$^{4}$, \\Heimo Krajnz$^{5}$ and Gerald Steinbauer-Wagner$^{1}$
\thanks{*This work was partially supported by the Austrian Research Promotion Agency (FFG) with the project EASIER (FO999886357).}
\thanks{$^{1}$Laurent Frering, Matthias Eder and Gerald Steinbauer-Wagner are with the Institute of Software Technology, Graz University of Technology, Graz, Austria. 
	 {\tt\small \{laurent.frering, matthias.eder, steinbauer\}@ist.tugraz.at}}%
\thanks{$^{2}$Bettina Kubicek and Dietrich Albert are with the Institute of Psychology, University of Graz, Graz, Austria. 
 	{\tt\small \{bettina.kubicek, dietrich.albert\}@uni-graz.at}}%
\thanks{$^{3}$ Denis Kalkofen is with the Institute of Computer Graphics and Vision, Graz University of Technology, Graz, Austria. 
 	{\tt\small kalkofen@icg.tugraz.at}}%
\thanks{$^{4}$ Thomas Gschwandtner is with Rosenbauer International AG, Linz, Austria. 
 	{\tt\small thomas.gschwandtner@rosenbauer.com}}%
\thanks{$^{5}$ Heimo Krajnz is with Professional Fire Brigade Graz, Graz, Austria. 
 	{\tt\small heimo.krajnz@stadt.graz.at}}%
}

\begin{document}

\maketitle
\thispagestyle{empty}
\pagestyle{empty}

\begin{abstract}


This paper presents a conceptual overview of the EASIER project and its scope. EASIER focuses on supporting emergency forces in disaster response scenarios with a semi-autonomous mobile manipulator. Specifically, we examine the operator's trust in the system and his/her cognitive load generated by its use. We plan to address different research topics, exploring how shared autonomy, interaction design, and transparency relate to trust and cognitive load. Another goal is to develop non-invasive methods to continuously measure trust and cognitive load in the context of disaster response using a multilevel approach. This project is conducted by multiple academic partners specializing in artificial intelligence, interaction design, and psychology, as well as an industrial partner for disaster response equipment and end-users for framing the project and the experiments in real use-cases.

\end{abstract}

\section{INTRODUCTION}


Assistance robots are useful tools to support emergency forces in disaster response situations. The robots can be used to collect information, operate fixtures or manipulate dangerous objects. However, fully autonomous systems are currently not feasible, both technically and due to a lack of acceptance by emergency forces \cite{SethuVijayakumar.2019}. There are multiple reasons for this: firstly, responders must have an appropriate level of trust in their equipment, which needs to be highly reliable in the field. Secondly, automated skills are currently still far from human-like performance in terms of complexity and reliability. Thus, mainly fully remotely-operated systems with limited skills are used in the field \cite{murphy14}. The main goal of using such robots is to keep humans out of dangerous zones while still allowing them to perform their tasks.

\begin{figure}[ht!]
	\centering
	\includegraphics[width=0.95\linewidth]{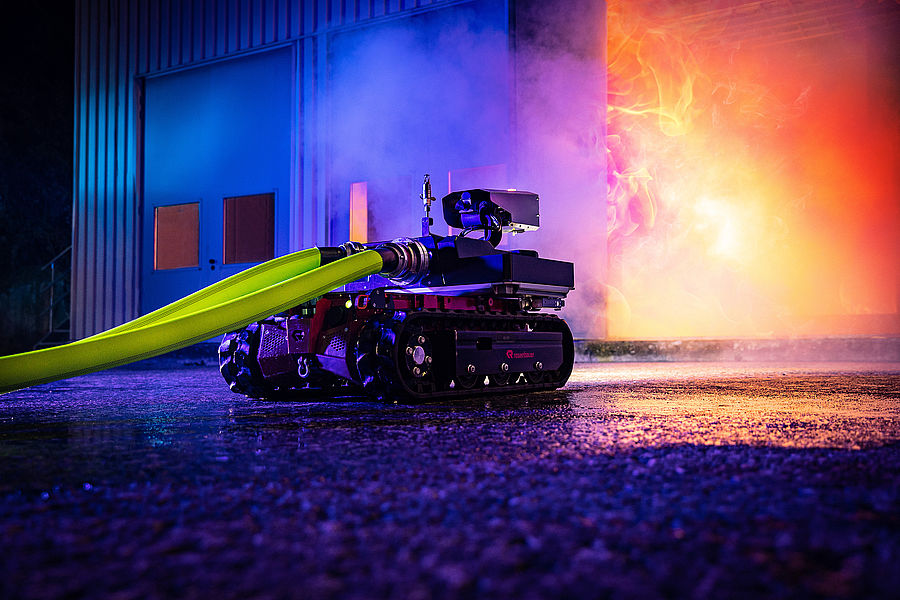} 
	\caption{The used mobile platform in a firefighting situation. An additional manipulator is developed for the project, which is placed on top of the mobile base.}
	\label{fig:disaster}
\end{figure}

To increase the use of assistance robots among first responders, an interdisciplinary team of roboticists, psychologists, user-interface specialists, firefighters, and a renewed vendor of firefighting gears work on mitigating the above drawbacks through an advanced Shared Autonomy concept and a robust non-invasive assessment of trust and cognitive load as part of the research project "Enabling and Assessing Trust when Cooperating with Robots in Disaster Response" (EASIER). The addressed use case which sets the frame for the project is mobile manipulation in disaster response. An industrial-grade mobile manipulator is used to conduct disaster-related tasks such as reconnaissance of hazard areas, clearing of entrances from debris, operation of valves and handles, or manipulation of dangerous goods. The research is conducted in close cooperation with active firefighters to identify relevant user requirements and conduct user studies in realistic disaster environments, answering a recent call for more field studies in HRI (Human-Robot Interaction) research \cite{weiss2021cobots}. Figure \ref{fig:disaster} shows an example firefighting situation using the robot platform.

\begin{figure}[ht!]
	\centering
	\includegraphics[width=0.95\linewidth]{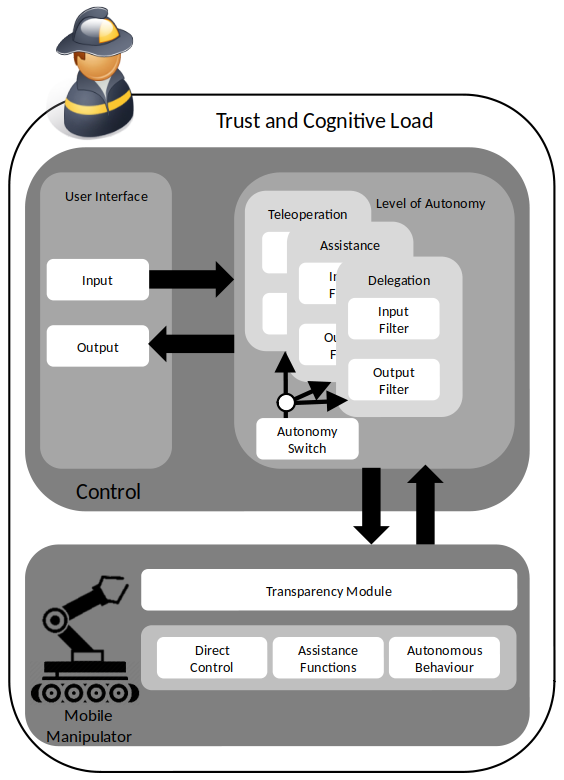} 
	\caption{Conceptual overview of the proposed Human-Robot Interaction concept in the area of disaster response.}
	\label{fig:overview}
\end{figure}

The research pursues two closely related goals. The first goal is to capture trust in the robotic system and the cognitive load of using the system, as both are crucial factors for human-robot interaction \cite{Ahmad.2019}. To assess trust and cognitive load without disturbing the operator during operation, non-invasive methods need to be developed, preferably through a multilevel approach that combines the operator's subjective evaluation with existing information from the robotic system, the user interface, and information about the user's interaction with them (e.g. reaction time, gaze). The need for using non-invasive methods results from the fact that mitigating a disaster is a very demanding task and any additional intervention compromises the realistic interaction. The second goal is to develop an advanced interaction concept based on the triangle of user interface, shared autonomy level, and transparency. A general overview of the concept is shown in Figure \ref{fig:overview}.

Using a valid non-invasive measurement technique we can investigate the effects of variations in the user interface, the shared autonomy level, and the transparency on trust and cognitive load. For the former, we will adapt the input and output of the user interface according to the actual task, situation, as well as estimated trust and cognitive load; these adaptations may happen offline or online and comprise abstraction or enhancement of the visualization or changing the control interface. The latter two both relate to Shared Autonomy as we aim at an optimized and transparent sharing of the responsibility and task. Based on the trust of the user in the robot's skills, the level of autonomy can be adapted from full teleoperation over assistant functions (similar to driver assistance systems) to full delegation of tasks. We plan to increase transparency to support the operator's situational awareness using the situation awareness-based agent transparency (SAT) model \cite{chen2014situation}. Based on the SAT model, we structure the information to enhance perception, comprehension, and projection of the robot's state, capabilities, and decision-making processes, which aims to increase the operator's trust in the robot without increasing cognitive load. 

Moreover, we envision that the robot is also able to estimate its reliability and certainty level for various situations to reduce the autonomy level in case of unforeseen circumstances the robot is not able to handle. Such an adaptation builds trust without increasing cognitive load only if transparent explanations are given by the robot. Such an interaction schema is rather novel in the area of disaster response.



In sum, the main research questions of the presented project are the following: 1) How can trust and cognitive load be measured in a non-invasive way and their importance be evaluated in the context of disaster response? and 2) How can interaction design, shared autonomy, and transparency work together to improve trust and reduce cognitive load in disaster response scenarios? 

The remainder of this paper is structured as follows. First, related work on the topic of shared autonomy, recent projects on disaster response, trust and cognitive load in robotic systems is presented. Then, the conceptual overview of the system design for answering the research questions is presented. Furthermore, integration details on the system concept are discussed and the next steps for the implementation of the proposed concept are presented.


\begin{figure*}
	\centering
	\includegraphics[width=0.8\textwidth]{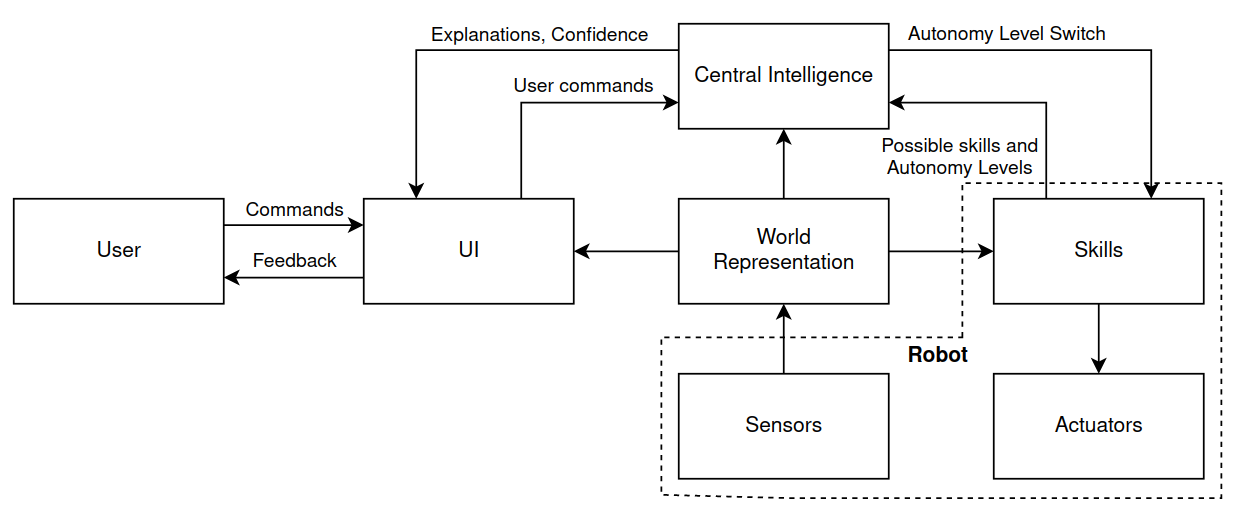} 
	\caption{System design highlighting the different interconnected components.}
	\label{fig:system}
\end{figure*}

\section{RELATED WORK}

This section discusses related work and developments regarding shared autonomy, as well as trust and cognitive load in robot operation for disaster response.

\subsection{Related Projects}
Multiple recent projects have tackled the problem of human-robot collaboration in disaster response.

NIFTi \cite{kruijff2014designing} was a project with close interaction with fire brigades, developing adaptive robots and interfaces taking into account the cognitive load of the users. The project notably experimented on multiple interfaces for mobile first responders \cite{Larochelle.03.11.201307.11.2013}. Building on NIFTi, TRADR \cite{kruijff2015tradr} focused on persistent models for perception, action, and collaboration in long-term missions.
SHERPA \cite{marconi2012sherpa} was a project with goals relating to multimodal HRI and decision making in the context of a multi-robot team commanded by one human in alpine scenarios where the human could be more or less available depending on the situation. Within S-HELP \cite{steiner2017psychological} a psychological framework to enable effective cognitive processing in the design of emergency management information systems (EMIS) focusing on decision-making processes, perception and information processing as well as personal and contextual variables was established. 
In the current project TRUST ME \cite{kubicek2019trust}, an instrument for measuring individual trust in robots will be developed and validated.

While similar in scope and research areas, these projects did not approach the issue of assistance robots for disaster response by focusing on how trust and cognitive load can be assessed using non-invasive methods and how the HRI can be improved by investigating the effects of different configuration of the level of autonomy, the interface design, and transparency.

\subsection{Shared Autonomy}
While shared autonomy gained increasing research interest in recent years, different definitions exist which define the term only roughly (see e.g. \cite{C.Brooks.2019,S.Hart.2016}). While some definitions focus on the technical aspects \cite{Nikolaidis.2017}, others highlight the importance of Human-Robot Interaction \cite{C.Brooks.2019}. 
As a result, the focus in this area relates to (combined) aspects such as the optimization of user input \cite{Nikolaidis.2017}, the recognition of human/robot intentions \cite{Javdani.2018}, the optimization of task distribution \cite{S.Hart.2016}, the monitoring of automated systems \cite{Marion.2018}, or the general optimization of interaction \cite{S.J.Anderson.2014}. Additionally, research to design interfaces to mitigate performance issues in robot teleoperation \cite{chen2007human} is carried out.
While all these definitions and fields of work do justify the term of shared autonomy, we see the term of shared autonomy in the presented project as a general interdisciplinary mean to improve assistance robots with a focus on aspects such as interface optimization, switching the level of assistant functionality, and transparent human/robot cooperation.

\subsection{Trust and Cognitive Load in Robot Operation}
The operator's trust in a robot system and the cognitive load caused by using the system are decisive for the acceptance of an assistance system by rescue teams \cite{Groom.2007}. Both, the level of trust and cognitive load depend on human-related, robot-related, and contextual factors 
\cite{Ahmad.2019}. For example, user studies showed a positive association between the level of transparency and the overall trust in the system \cite{Boyce.2015}. Moreover, the robot's reliability affects the operator's trust \cite{Desai.2012} and trust repair \cite{Baker.2018}. Operators’ trust and workload are also influenced by the level of autonomy \cite{onnasch2014human}.
It is also shown that explanations regarding robots behaviour can increase trust in robots \cite{wang2018my}.

Using a non-invasive multilevel approach including subjective data, psychophysiological measurements, performance variables, reconstructional video, and observation techniques, we will assess trust and cognitive load without interfering with the operator during the current mission \cite{Colin.2014}.


\section{System Overview}

One important aspect of the project is the design and construction of the robot. This is done in close collaboration with an industrial partner experienced in robotic devices for assistance in firefighting, who will provide the mobile base and the robotic arm and integrate the required sensors. The robot will have two main skills, the navigation of the mobile platform and the control of the robotic arm to manipulate objects in the environment.
For navigation, the robot will be equipped with two LiDARs and with 2 camera arrays composed of a stereo camera and a depth camera to increase the situational awareness of the robot and to improve the robot operation. One camera array will be attached to the front of the robot and one attached to the arm, to provide visual feedback for the autonomy module as well as to the operator. The cameras are to be used for object detection for save navigation and manipulation, both either when using teleoperation or in the assistance mode with shared autonomy. Furthermore, a 360-degree stereo camera array will be mounted on the back of the robot to improve the operator's situational awareness during teleoperation.

As the necessary backbone to start answering the aforementioned research questions, the system design revolves around 4 components: 1) the user interface (UI) providing a continuous flux of information to the user, 2) the robot mentioned above allowing for dynamically switching between levels of autonomy for multiple skills 3) the central intelligence computing reliability and certainty levels for different skills, triggering autonomy level switches and UI adaptation, and 4) the sensors creating a world representation. An overview is given in Figure \ref{fig:system}.

The main workflow is that given a certain state of the world, its observation leads to a world representation that allows for skills to be performed at specific levels of autonomy with estimated reliability and certainty levels. This information is reflected on the user interface, and the user inputs determine the specific skill and autonomy level to be used. The UI then provides a specific interface tailored for each skill and autonomy level.

During skill execution, the central intelligence also updates a confidence metric for the given task. This metric is defined task by task, for example by estimating a collision probability during navigation \cite{axelrod2018provably}. This can be used both to better inform the user and to trigger or advise for autonomy level switches, for example when a failure is detected or the task is too complex for the robot. In parallel to this, explanations for these decisions are also computed for transparency purposes.

As a first approach, we focus on two different autonomy levels, namely teleoperation and semi-autonomous operation. Teleoperation will be supported by assistance functions (e.g. inverse kinematics, obstacle avoidance) and semi-autonomous operation will make use of different tools to support robot skills (e.g. waypoint navigation, grasping target selection).

The available skills and their specific interaction functionalities are reflected on the user interface alongside reliability and certainty information and a world representation devised from the sensors. The interface is designed to be used either using traditional displays or virtual reality, the design process is done alongside end-users to guarantee the best possible applicability of the system.

Explanations for the decisions made by the central intelligence (e.g. triggering an autonomy level switch, estimating reliability and certainty for a task) will also be computed and displayed. One of the goals for future experiments is to have a better understanding of the amount and content of the explanations to guarantee enough transparency to build trust without impairing the usability of the system due to information overload.

The presented concept aims to develop psychological methods and models to determine the operator's trust and cognitive load, focusing on a non-invasive multilevel approach to not impair the control and communication with the robot in case of a disaster by unavoidable disturbances.

This project is done in cooperation with end-users, with a continuous feedback loop to improve the methodology and the individual components.


\section{Current work and Future Plans}

As a first step to improve the HRI, a fast method to provide detailed explanations of failures in robot motion planning was proposed in \cite{Eder.2022}. By identifying failures origin from planning constraints, possible failure explanations can be provided in logarithmic time, also for large constraint sets. A study to investigate the impact on HRI was also conducted on a research platform and shows significant improvements in trust and task performance when explanations of the errors are provided.

In the current state of the project, the next goal is to finish the integration of the mobile manipulator with its sensors and implement basic navigation and manipulation skills, as well as providing a first iteration of the interaction design. Moreover, the concept of the central intelligence and its interaction with the user, the world model, and the skills needs to be refined.

The software development will be fully ROS-compatible and includes a simulated environment to facilitate experiment design, and in a second stage help in investigating assistance functions and skill-related UI.

Another experiment is planned to measure trust and cognitive load in a navigation task with varying difficulty (e.g. faulty sensors, different obstacle setups) and autonomy level switch between teleoperation and semi-autonomous navigation. We aim at integrating an estimation of motion planning reliability and certainty for informing the user and triggering autonomy level switches. The previously mentioned robot sensors will be used, as well as an overall scene camera and cameras pointing towards the user for further analysis.

For estimating the operator's level of trust and cognitive load using non-invasive methods, data collection studies will be conducted in these experiments with experienced rescue operators as test participants. Research question 1) will be addressed using data collected during the experiments and with posthoc reconstruction. To answer research question 2) user studies following a multi-level approach to investigate the influences of the defined levers are to be conducted.


\section{Conclusion}
This paper presented the current scope and plans for the EASIER project. The main objective is to integrate shared autonomy, interaction design, and transparency to provide a trustworthy system for disaster response scenarios while limiting cognitive load when using said system. Multiple developments are expected, from robot design to a better understanding of how trust and cognitive load can be measured and improved on in the context of disaster response. To this end, we specifically plan to explore topics relating to shared autonomy, interaction design, and transparency. The ultimate goal is to validate these developments in real-world scenarios designed directly with end users.








\bibliographystyle{IEEEtran}
\bibliography{IEEEabrv,IEEEexample}

\end{document}